\documentclass[final]{nesy2026} 

\usepackage{longtable}

\usepackage{booktabs}
\usepackage[load-configurations=version-1]{siunitx} 

\newcolumntype{M}[1]{>{\centering\arraybackslash}m{#1}}
\usepackage{multirow}
\usepackage{algorithm}
\usepackage{algcompatible}
\usepackage{listings}
\usepackage{adjustbox}
\usepackage{fancyvrb}

\theorembodyfont{\upshape}
\theoremheaderfont{\scshape}
\theorempostheader{:}
\theoremsep{\newline}

\title[Assembly Learning Under Semantic Constraints]{Learning to Assemble Novel Structures with Unfamiliar Parts under Semantic Constraints}

 \clearauthor{\Name{Jonghyuk Park} \Email{jay.jh.park@ed.ac.uk}\\
  \Name{Alex Lascarides} \Email{alex@inf.ed.ac.uk}\\
  \Name{Subramanian Ramamoorthy} \Email{s.ramamoorthy@ed.ac.uk}\\
  \addr School of Informatics, University of Edinburgh \\
  10 Crichton Street, Edinburgh EH8 9AB, UK}

\begin{document}

\maketitle

\begin{abstract}
This paper describes a neurosymbolic architecture for learning to assemble novel structures using evidence from embodied conversations and task demonstrations.
We focus on scenarios where an agent encounters, after deployment, \textit{semantic constraints} on structures---in other words, constraints as to which part types and features make valid structures---that were not available during training, and where it is \textit{initially unaware} of the relevant structure and component part concepts.
The agent must acquire and exploit such knowledge through user interactions \textit{while} attempting assembly.
We study this setting in a simulated toy truck assembly domain, learning from symbolic evidence encoded in natural language and from dense visual observations.
Our experiments show that communicating semantic constraints through natural language (e.g., ``dump trucks have a dumper'') yields more data-efficient online adaptation than relying only on task demonstrations and/or only naming the parts through natural language.
\end{abstract}

\section{Introduction}
\label{sec:intro}

Robotic assembly in open-ended environments requires agents to cope with task knowledge that may not be available before deployment \citep{jiang2022review}.
Prior work commonly distinguishes assembly knowledge into i) geometry of parts and joins, ii) assembly orders, and iii) low-level control skills \citep{lee2024human}.
These aspects comprise largely physical constraints and are well suited to learning from task demonstrations \citep{zhu2018robot}.

\begin{figure}[t]
\floatconts
    {fig:intro}
    {\caption{Illustration of an assembly domain, and an example user-agent interaction.}}
    {
        \subfigure[\small Truck types defined by parts.]{
            \label{fig:truck_domain}
            \includegraphics[width=0.35\textwidth]{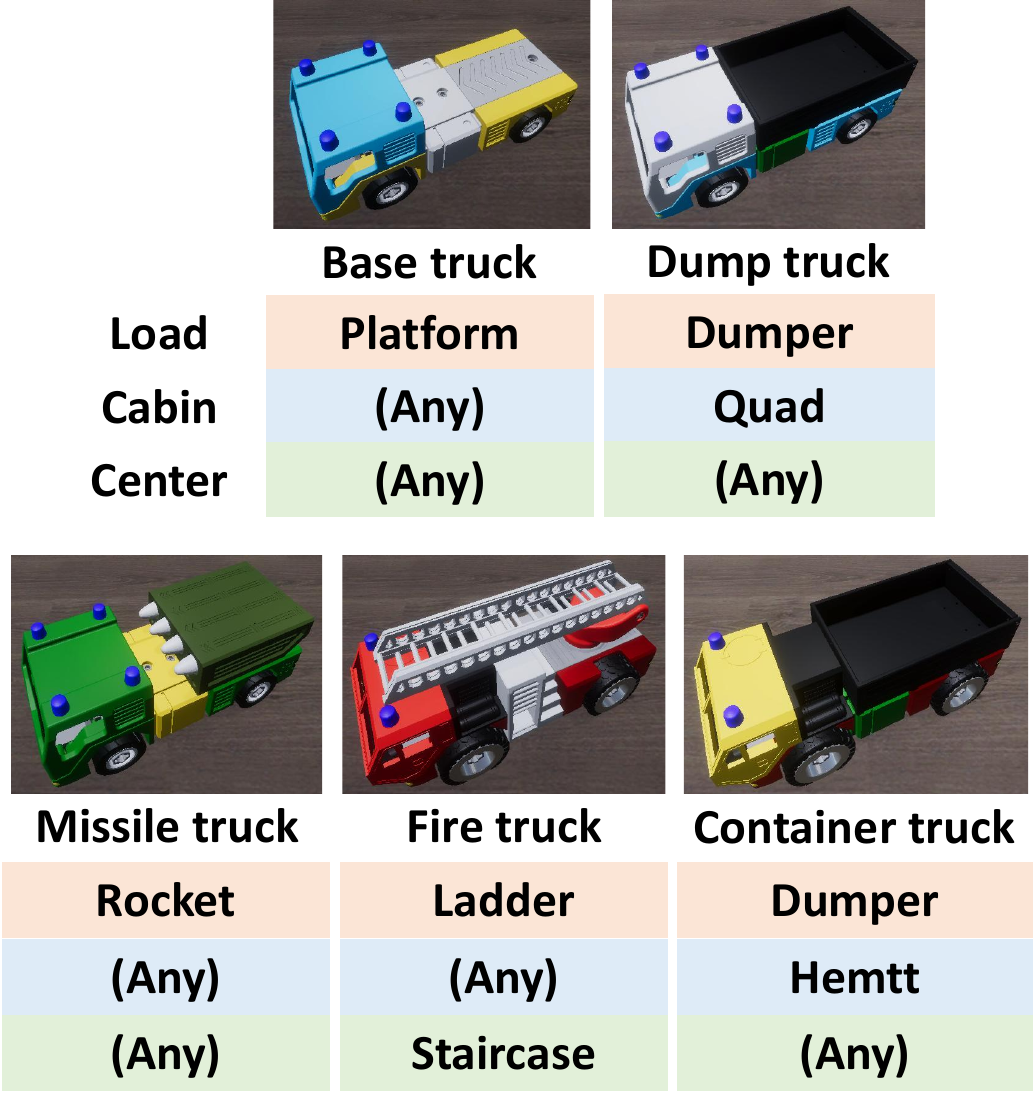}
        }\qquad\qquad
        \subfigure[\small User-agent dialogue.]{
            \label{fig:opening_ex}
            \includegraphics[width=0.275\textwidth]{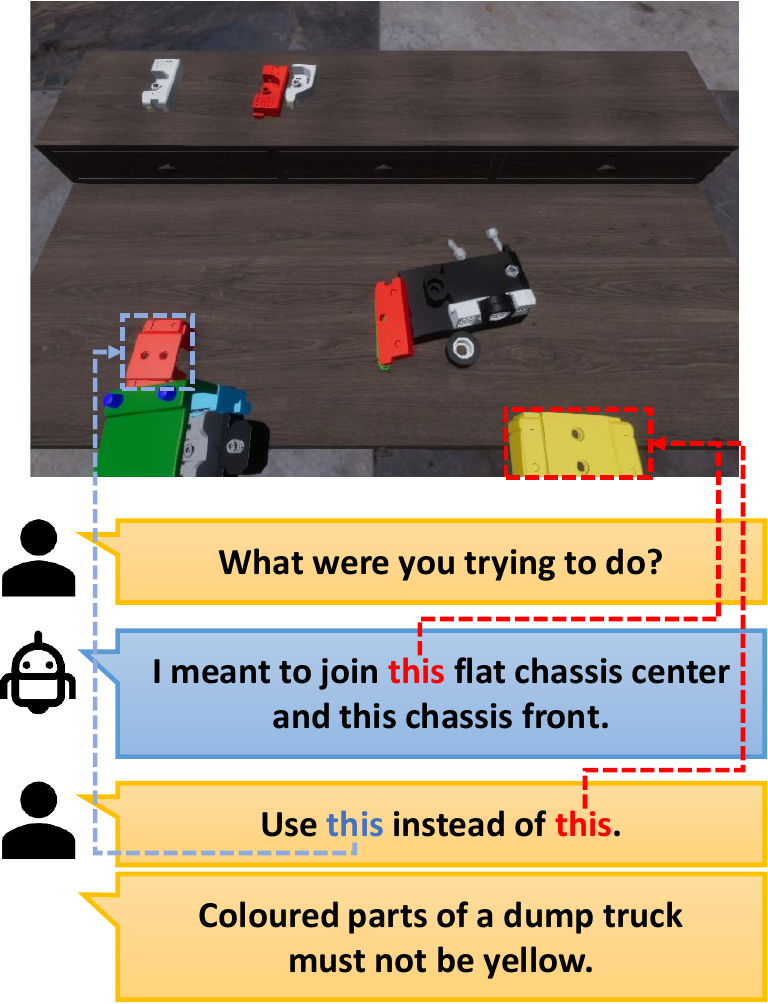}
        }
    }
\end{figure}

However, successful assembly may also depend on \textit{semantic constraints}: arbitrary domain conventions under which it is physically feasible to join two parts, but doing so creates an invalid structure.
In the example shown in Fig.~\ref{fig:opening_ex}, the join attempt is corrected despite geometric compatibility because it violates a semantic rule: namely that ``Coloured parts of a dump truck must not be yellow".
Demonstrations alone are insufficient as a corrective signal here, since multiple rules may be consistent with them: the constraint in Fig.~\ref{fig:opening_ex} might have been, for example, ``Dump trucks must have a red chassis center''.
The scope of the rule is also latent, as it may apply to a specific subtype of truck or to all trucks.
The challenge becomes even greater when the agent is initially unaware of the concepts in which the constraints are expressed; e.g., if it never encountered a flat chassis center nor its label during training.
The agent cannot logically infer a rule referring to a concept that it doesn't know exists.

We address this setting with a neurosymbolic agent architecture for online adaptation from embodied natural language (NL) interaction and task demonstrations.
The central challenge is that domain knowledge may be introduced piecemeal during deployment, and the agent must change its behaviour on the timescale of the current interaction.
This motivates separating dense perceptual grounding from explicit symbolic memory and planning: visual classifiers can be updated via new exemplars; the lexicon and knowledge base can be extended with new concepts and constraints; and the planner can immediately use the updated knowledge to revise object selection and action sequences.
The neurosymbolic coupling is therefore primarily perception-to-reasoning: neural perception supplies graded candidate groundings, and newly acquired symbolic knowledge selects among these groundings and revises action plans, without directly revising the visual feature representation.
This design targets the online adaptation problem studied here, while leaving tighter symbolic-to-neural feedback as a compatible extension for settings that require representation-level revision.


We evaluate the proposed framework in a simulated toy truck assembly domain.
Truck types are defined by constraints over part subtypes and attributes (see Fig.~\ref{fig:truck_domain}).
We compare an agent that can interpret generic NL statements expressing semantic constraints---such as \textit{dump trucks have a dumper}---against ablative baselines that receive demonstrations, part labels, and/or non-generic correction signals.
The results show that communicating semantic constraints through NL generics improves data efficiency for online learning in this assembly domain, beyond the benefits of naming novel part concepts alone.


We publicly release the codebase implementing our agent architecture and the simulation environment for experiments at \url{https://github.com/jpstyle/semantic-assembler}.


\section{Semantic Assembly: Task Setting and Domain}
\label{sec:semantic_assembly}

\subsection{Task Domain, Goal and Input Formulation}

In our formulation, a semantic assembly domain defines two sets of concepts, $\Pi$ and $\Sigma$:
\begin{itemize}
    \item $\Pi$ is a set of \textit{atomic part concepts}, represented as unary predicates over primitive objects; their instances are minimal building blocks of assembly structures: e.g., \textsf{dumper}, \textsf{quad\_cabin}, \textsf{cabin}.
    \item $\Sigma$ is a set of \textit{subassembly concepts}, represented as unary predicates over composite structures; their instances consist of more than one atomic parts assembled together: e.g., \textsf{truck}, \textsf{fire\_truck}, \textsf{truck\_front}.
\end{itemize}
Hyper/hyponymy, or \textit{is-a}, relations are represented as subclass implications between unary predicates in $\Pi$ and in $\Sigma$:
for instance, \textsf{quad\_cabin} is a subtype of \textsf{cabin}; and 
\textsf{fire\_truck} is a subtype of \textsf{truck}.
Holo/meronymy, or \textit{has-a}, relations are represented using the binary predicate $have(x,y)$ between $\Pi$ and $\Sigma$ (e.g., A \textsf{truck} has a \textsf{cabin}) or within $\Sigma$ (e.g., A \textsf{truck} has a \textsf{truck\_front}).
Together, these predicates and implication/relation rules constitute the domain theory $\Omega$.

Each instance of a semantic assembly task is characterised by: (a) two sets of objects $C$ and $D$ laid out on the tabletop; and (b) an assembly goal $\gamma\in\Sigma$.
$C$ contains exactly the `ground-truth' atomic parts required to build an instance of $\gamma$, while $D$ is a (possibly empty) set of `distractor' parts, which are geometrically feasible for assembly but will violate a semantic constraint if used.
Given the subtype rules within $\Pi$, an object in $C$ or $D$ may satisfy multiple part predicates.
Parts also have colour attributes, and so may be referred to as `a red quad cabin' in NL, for example.

Each goal concept $\gamma\in\Sigma$ is associated with two formal specifications $G_\gamma$ and $S_\gamma$, which together define the range of valid assembly structures admissible as an instance of $\gamma$:
\begin{itemize}
    \item $G_\gamma=(V_\gamma,E_\gamma)$ is an intensional structural specification (`assembly topology graph') of $\gamma$ that specifies the necessary and sufficient part joins in a valid instance of $\gamma$.
    Each node in $V_\gamma$ specifies a type requirement from $\Pi$ or $\Sigma$.
    The edges $E_\gamma$ specify required joins and their relative poses.
    \item $S_\gamma$ is a set of quantified first-order logic (FOL) rules or integrity constraints relevant to $\gamma$ (see Fig.~\ref{fig:truck_domain}), which must hold in all instances of $\gamma$.
    For instance, ``Dump trucks have a dumper" is encoded as $\forall x\exists y.dumpTruck(x)\rightarrow dumper(y)\land have(x,y)$.
\end{itemize}
The task objective is to plan and execute a sequence of actions that constructs a structure satisfying $G_\gamma$ while observing all constraints in $S_\gamma$, using only parts from $C$ and not $D$.
The agent has a single-view RGB image $\mathcal{I}\in[0,1]^{3\times H\times W}$ of the scene, where $H$ and $W$ are the sensor height and width respectively.
During assembly, the agent engages in an NL dialogue with a teacher who supervises the task execution; see \S\ref{sec:semantic_assembly:user_interactions} for details.

\subsection{Scope and Assumptions}
\label{sec:semantic_assembly:assumptions}

In this study, we make several simplifying assumptions to isolate deployment-time acquisition of semantic assembly knowledge from low-level robotics and open-domain language-understanding challenges.
Experiments are conducted in a simulated toy-truck domain and we evaluate online adaptation within this domain, rather than cross-domain transfer or real-world robustness.
The agent receives oracle object masks and 6D poses from the simulator.
Demonstrations are assumed to be segmented into parametrised primitive actions which are executed by an oracle controller.
Teacher interactions follow a controlled dialogue protocol, so the experiments test the effect of receiving generic semantic content rather than open-domain NL understanding.
The visual feature backbone is fixed: online perceptual adaptation updates exemplar memories and concept classifiers, not visual features.
Colour predicates are grounded by the same exemplar-based procedure as part concepts, but their classifiers are initialised before the main assembly episodes from a small set of positive and negative exemplars.
Finally, instances of the same atomic part concept are assumed to share a 3D geometry, enabling type-level point-cloud and join-pose representations.

\subsection{Initial Agent Knowledge and Task Abstraction}

We consider a setting in which the agent possesses domain-neutral skills but lacks domain-specific assembly knowledge.
The domain-neutral skills include primitive action interfaces and a high-level symbolic planning faculty that deploys them.
We adopt {\em Answer Set Programming} (ASP; \citealt{lifschitz2019answer}), a declarative programming approach based on normal logic programs.
We use ASP because it allows seamless integration of learned semantic constraints in planning, also encoded as ASP program fragments, thanks to its ability to model indirect action effects \citep{tran2023answer}.
App.~\ref{app:asp_encoding} describes how the assembly planning problem is implemented as ASP programs.

At the start, the agent is ignorant of domain-specific knowledge: the concept sets $\Pi$ and $\Sigma$ (i.e., the hypothesis space of possible parts and structures), the domain theory $\Omega$, the 2D-visual and 3D-geometric features of the atomic parts in $\Pi$, the assembly topology graphs $G_\gamma$ and the semantic constraints $S_\gamma$ for each $\gamma\in\Sigma$ are empty.
The agent thus lacks the hypothesis space of possibilities and must acquire it after deployment.
Our main hypothesis is that generic NL statements improve online learning by communicating reusable semantic constraints, beyond what can be obtained from demonstrations and part labels alone.

\subsection{Agent-Teacher Interactions}
\label{sec:semantic_assembly:user_interactions}

Task demonstrations and NL dialogue provide complementary evidence.
Demonstrations convey continuous information, in particular the relative poses required to join parts.
Within our controlled dialogue protocol, each episode begins with a teacher-specified goal: ``Build a $\gamma$".
If the agent lacks the target concept or required part concepts, the teacher provides either a demonstration, a verbal definition, or labelled exemplars.
During planning and execution, agent failures expose what kind of knowledge is missing: an ``Is there a dumper?'' query calls for a labelled exemplar; a mistaken ``I meant to join this quad cabin'' response lets the teacher correct the part type; and a geometrically feasible but invalid join calls for a semantic constraint.
Teacher feedback targets the exposed gap, updating visual exemplars, the lexicon, or symbolic memory before replanning.
A fuller account of the possible interaction flows, along with a visual flowchart depiction, is provided in App.~\ref{app:interaction_flow}.


\section{The Neurosymbolic Agent Architecture}

This section describes the neurosymbolic architecture used to implement the interaction protocol above.
Before describing individual modules, Alg.~\ref{alg:episode} summarises the episode-level control loop: perception produces candidate symbolic groundings, dialogue updates memory when knowledge gaps are exposed, symbolic reasoning selects goals and plans joins, and actuation executes the selected actions.
Fig.~\ref{fig:arch} illustrates the overall architecture.

\begin{algorithm}[t]
\footnotesize
\caption{Interactive semantic assembly episode}
\label{alg:episode}
\begin{algorithmic}[1]
\REQUIRE goal utterance $u$, scene image $\mathcal{I}$, visual exemplar base $XB$, symbolic knowledge base $KB=(\Omega,\{G_\sigma,S_\sigma\}_{\sigma\in\Sigma})$, lexicon $L$
\STATE Map $u$ to target concept $\gamma$ using lexicon $L$.
\IF{$\gamma$ or required concepts are unknown}
    \STATE Obtain teacher definition, labelled exemplars, or demonstration.
    \STATE Update $XB$, $KB$, $L$ as applicable.
\ENDIF
\WHILE{task not accepted by teacher}
    \STATE Perceive scene from $\mathcal{I}$ and produce candidate part groundings with scores.
    \STATE Solve goal-selection ASP to choose an object-filled goal structure under current $G_\gamma$, $S_\gamma$, and grounding.
    \IF{a required part cannot be grounded}
        \STATE Ask teacher for an instance and update $XB$.
        \STATE \textbf{continue}
    \ENDIF
    \STATE Solve join-sequence ASP using the selected goal structure, then execute the resulting sequence.
    \IF{teacher interrupts}
        \STATE Explain intended action.
        \STATE Update $XB$ and/or $KB$ according to the diagnosed error and active strategy.
    \ENDIF
\ENDWHILE
\end{algorithmic}
\end{algorithm}

\subsection{Vision Processing Module}
\label{sec:architecture:vision_processing}

The vision processing module maps dense visual inputs to structured symbolic representations consumed by the planning and learning components.
Given a single-view RGB image of the scene, the module outputs, for each detected object instance: 1) a set of candidate atomic part concepts from $\Pi$ with associated confidence scores; and 2) a 3D geometric representation suitable for specifying join relations.
Confidence scores produced by the vision module are propagated to the symbolic layer, where they influence planning decisions.
Implementation details of feature extraction, classification, geometry extraction, and pose handling are given in App.~\ref{app:vision_module}.

\begin{figure}[t]
\floatconts
    {fig:arch}
    {\caption{Overview of the neurosymbolic agent architecture.}}
    {\includegraphics[width=0.875\textwidth]{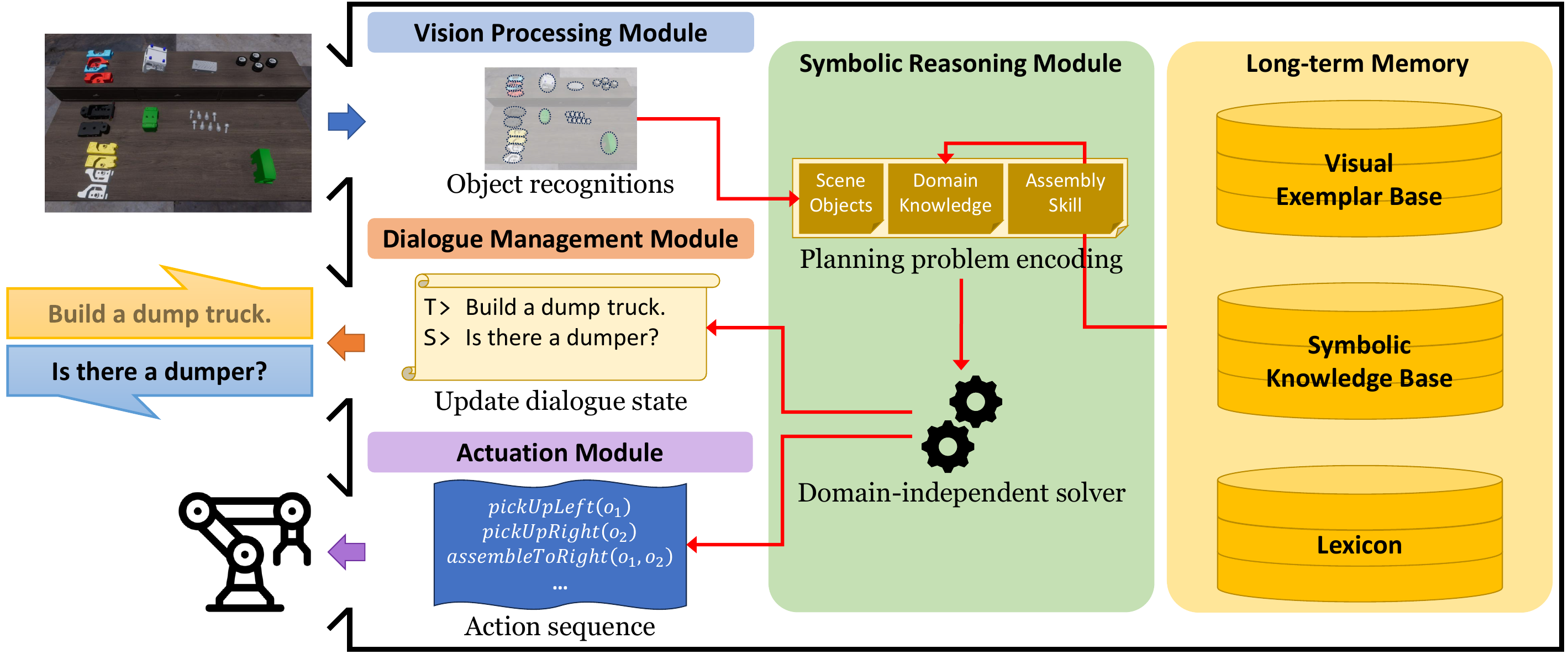}}
\end{figure}

\subsection{Dialogue Management Module}
\label{sec:architecture:dialogue_management}

The dialogue management module interprets teacher utterances and generates agent responses in the controlled interaction protocol of \S\ref{sec:semantic_assembly:user_interactions}.
It handles goal utterances, clarification requests, corrective feedback, and generic statements that express semantic constraints.
Our implementation deploys an off-the-shelf large-coverage semantic parser \citep{copestake2000open}, appended with a heuristic postprocessing pipeline.
Here, generic statements that express semantic constraints take two forms:
\begin{itemize}
    \item Universal constraints dictate that a part type of a subassembly must all have certain qualities.  For example, ``All fenders of fire trucks are red" is:
    \begin{align*}
    \forall x\forall y.fireTruck(x)\land fender(y)\land have(x,y)\rightarrow red(y)
    \end{align*}
    \item Existential constraints, that a certain subassembly must have at least one part with certain qualities.
    For example, ``Missile trucks have a green fender.":
    \begin{align*}
    \forall x\exists y.missileTruck(x)\rightarrow fender(y)\land green(y)\land have(x,y)
    \end{align*}
\end{itemize}
These FOL statements undergo an additional automated translation step into ASP program clauses, to be integrated with ASP planning problems.
The above examples become the following, where $c_i$ is the index of the violated constraint:
\begin{lstlisting}[frame=single,linewidth=\textwidth,mathescape=true,numbers=none,aboveskip=1.5mm,belowskip=1.5mm]
$violated(c_i)\leftarrow fireTruck(X),fender(Y),have(X,Y),\text{ not }red(Y).$
\end{lstlisting}
\begin{lstlisting}[frame=single,linewidth=\textwidth,mathescape=true,numbers=none,aboveskip=1.5mm,belowskip=1.5mm]
$violated(c_i)\leftarrow
  missileTruck(X),\#count\{Y:fender(Y),green(Y),have(X,Y)\}=0$.
\end{lstlisting}

\subsection{Actuation Module}

The actuation module executes the primitive actions selected by the symbolic planner using the oracle simulator controller described in \S\ref{sec:semantic_assembly:assumptions}.

\subsection{Long-term Memory Module}

The long-term memory module stores the knowledge updated through Alg.~\ref{alg:episode}: the visual exemplar base (XB), the symbolic knowledge base (KB), and the lexicon.
The visual XB stores positive and negative visual exemplars for each atomic part concept $\pi\in\Pi$, respectively $\chi^+_\pi$ and $\chi^-_\pi$.
Each exemplar is represented as a feature vector generated by the vision processing module.
Whenever either $\chi^+_\pi$ or $\chi^-_\pi$ is updated, a new binary classifier for $\pi$ is induced.
New entries are added when the teacher says an object is (not) an example of $\pi$.

The symbolic KB stores the domain theory $\Omega$, the assembly topology graphs $G_\gamma$, and the semantic constraint sets $S_\gamma$ for each $\gamma\in\Sigma$.
As explained in \S\ref{sec:semantic_assembly:user_interactions} and \S\ref{sec:architecture:dialogue_management}, these structures are populated from teacher definitions, demonstrations, and generic utterances.

The lexicon maps concepts in $\Pi$, $\Sigma$ to their NL labels.
We do not directly use NL labels as the internal indices $\pi\in\Pi$; instead, the lexicon tracks associations between internal concept indices and teacher-provided labels.
This separation allows us to compare strategies with and without shared part vocabularies.

\subsection{Symbolic Reasoning Module}
\label{sec:architecture:symbolic_reasoning}

The symbolic reasoning module implements the two ASP planning subproblems in Alg.~\ref{alg:episode}: goal selection and join-sequence planning.
Long-horizon planning for assembly under semantic constraints has a combinatorially large search space, so we decompose the planning problem into the two subproblems.
In the goal selection phase, the agent selects the best set of recognised scene objects that can be used to assemble the goal concept $\gamma$ by building a structure instantiating the assembly topology $G_\gamma$.
Currently known constraints in $S_\gamma$ are translated into ASP clauses (see above) and added to the ASP encoding of the goal selection problem, so that the agent avoids knowingly violating the constraints.

Each possible goal structure is scored by the visual compatibility of part recognition, as measured by the sum of corresponding confidence scores, and the count of any violated constraints in $S_\gamma$.
The result of the goal selection process is that the agent makes decisions as to which objects are instances of which (currently known) atomic part concepts.
Our implementation uses the multi-shot solving feature of the ASP solver Clingo \citep{gebser2019multi} for incremental optimisation of the score.
In case the agent fails to visually recognise the full set of objects needed for building $\gamma$, our ASP encoding encourages partial planning towards fulfilling $G_\gamma$, after which the agent reports planning failure and requests feedback from the teacher (see the ``grounding failure" link in Fig.~\ref{fig:interaction_flow}).
Our encoding also accommodates re-planning during execution after planning failures or teacher corrections.

The output of the goal selection phase is passed to the join sequence planning phase, in which the agent plans a collision-free order of joins.
Clingo's capability to solve ASP programs modulo theories \citep{gebser2016theory} allows us to check whether there exists a collision-free joining path between two subassemblies by integration of any arbitrary motion planner.
Any infeasible joins are discarded and remembered \textit{during} plan search.
We record in App.~\ref{app:results_aux} unique motion-planner calls as an auxiliary metric, since expensive collision checks can affect performance.
The planned action sequence is finally passed to the actuation module for execution.


\section{Interactive Learning Procedures}
\label{sec:interactive_learning}

\subsection{Agent-Teacher Interaction Strategies}

We compare four interaction strategies that share the same architecture, task abstraction, visual backbone, planner, and oracle execution assumptions, but differ in what kinds of teacher feedback they can use to update memory.
\begin{description}
    \item [NoLabels+CaseMemory:] The agent does not share a NL vocabulary for the part concepts in $\Pi$.
    It cannot interpret concept labels or generic statements, but it can still store correction cases over its own internal concept symbols when teacher interventions indicate that one attempted choice should be replaced by another.
    This serves as a diagnostic lower bound for learning without shared part labels.
    \item [LabelsOnly:] The agent can use shared vocabularies obtained from NL dialogue for part concept labelling, but does not store semantic constraints from teacher corrections.
    \item [Labels+CaseMemory:] The agent can use shared part concept labels and also stores non-generic correction cases of the form ``Use $this_1$ instead of $this_2$'', but cannot interpret generic statements that encode constraints in $S_\gamma$.
    \item [Labels+TeacherRules:] This is our full approach.
    The agent can use concept labels and teacher-provided generic statements that are translated into symbolic constraints.
\end{description}

\subsection{Learning Visuals and Geometries of Parts}

The three label-aware strategies acquire labelled exemplars for part concepts directly through dialogue and narrated demonstrations, enabling targeted updates to visual grounding.
By contrast, NoLabels+CaseMemory lacks a shared part vocabulary and relies on indirect signals, including post-hoc analysis of completed assemblies and teacher interruptions, which introduces additional noise.
The extraction of 3D geometries and join poses follows the same procedure for all strategies; label-aware agents can immediately identify novel parts via linguistic cues, whereas NoLabels+CaseMemory relies on recognition uncertainty.
The detailed procedures for visual grounding and 3D geometry extraction are provided in App.~\ref{app:learning_vis}.

\subsection{Learning Semantic Constraints}

Labels+TeacherRules learns semantic constraints directly from generic teacher statements, as shown in Fig.~\ref{fig:interaction_flow}.
When the agent violates a semantic constraint, the teacher states the violated rule, and the agent translates it into FOL and ASP clauses added to $S_\gamma$.
LabelsOnly performs no such semantic-memory update.

The +CaseMemory strategies store non-generic correction memories derived from teacher interventions.
When a violation is signalled by ``Use $this_1$ instead of $this_2$'', the agent stores a most-specific correction case that discourages using $this_2$ (as recognised) when an unused $this_1$ (as recognised) is available.
For example, suppose the agent is tasked to build a dump truck and is notified to use a red hemtt cabin instead of a blue quad cabin.
The agent would then add a new ASP constraint as follows; informally, `when building a dump truck, do not use a blue quad cabin when a red hemtt cabin is available':
\begin{lstlisting}[frame=single,linewidth=\textwidth,mathescape=true,numbers=none,aboveskip=1.5mm,belowskip=1.5mm]
$violated(c_i)\leftarrow dumpTruck(X),quadCabin(Y),blue(Y),have(X,Y),$
  $\hspace*{1in}hemttCabin(Z),red(Z),\text{ not }have(X,Z).$
\end{lstlisting}
This case-memory mechanism does not abstract over corrections or discard irrelevant literals; inducing general constraints from cases, e.g. via ILP or anti-unification, is a natural direction for future work.


\section{Experiments}

\subsection{Experimental Design and Evaluation}

We test the interaction strategies on 30 datasets from the simulated toy-truck domain described in App.~\ref{app:domain_full}, each comprising 40 randomly sampled online assembly episodes.
The domain consists of 22 atomic part types in $\Pi$ and 6 truck types in $\Sigma$.
The first 5 problems in each dataset are a `warm-up', where the agent is tasked to assemble a generic truck without any distractors ($D=\varnothing$).
After this, unforeseen subtypes of trucks and parts are constantly introduced along with randomly sampled distractors ($D\neq\varnothing$).
There are no distinct `training' and `testing' splits---learning happens \textit{during} task execution and the interaction with the teacher.
The primary comparison is among the three label-aware strategies; NoLabels+CaseMemory is retained as a diagnostic lower bound for learning without shared part labels.

We use cumulative regret as the primary evaluation metric: i.e., the accumulated counts of errors made across a sequence of 40 planning problems.
We track five types of errors in total: 1) joining structurally incompatible pairs; 2) joining at incorrect pose; 3) using a distractor part; 4) failures to ground a needed part (addressed by ``Is there a X?" questions); and 5) ASP planner timeouts due to excessive grounding uncertainty.
These five error types are weighted uniformly.
We also monitor how the agents' visual grounding evolves over time, measured as mean F1 score curves obtained across each dataset.
We report the cumulative regret and mean F1 curves averaged over the 30 datasets with 95\% confidence intervals.

\subsection{Results and Discussion}

\begin{figure}[t]
\floatconts
    {fig:results}
    {\caption{Cumulative regret and mean F1 score curves with 95\% confidence intervals.}}
    {
        \subfigure[\small Cumulative regret curves]{
            \label{fig:cumul_regrets}
            \includegraphics[width=0.425\textwidth]{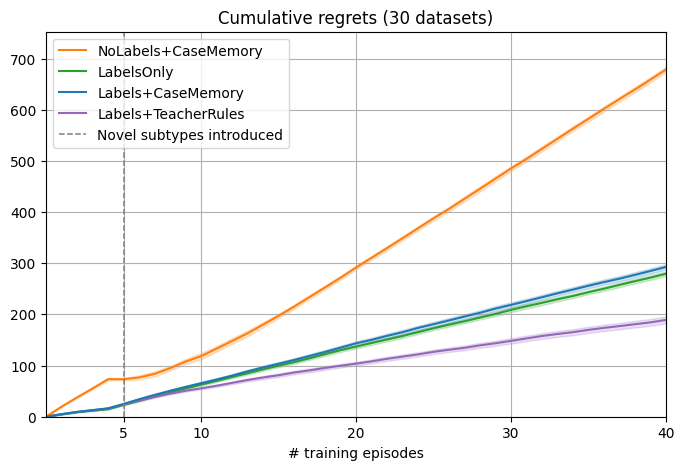}
        }\quad
        \subfigure[\small Mean F1 curves]{
            \label{fig:mean_f1s}
            \includegraphics[width=0.425\textwidth]{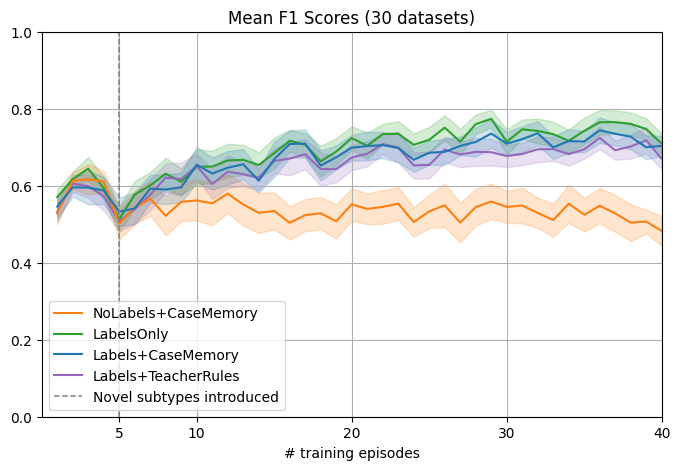}
        }
    }
\end{figure}

Fig.~\ref{fig:cumul_regrets} presents the averaged cumulative regret curves across 40 episodes for each interaction strategy.
NoLabels+CaseMemory incurs substantially higher regret than the label-aware strategies, reflecting the difficulty of learning novel part concepts without a shared vocabulary.
Among the label-aware strategies, LabelsOnly and Labels+CaseMemory exhibit very similar regret curves, with overlapping confidence intervals and no reliable advantage from storing most-specific correction cases.
This suggests that simple case memory is not an effective substitute for reusable semantic constraints in this domain.
By contrast, Labels+TeacherRules consistently achieves lower cumulative regret, showing that teacher-provided generic constraints improve online assembly learning beyond part labels and non-generic correction memories.
App.~\ref{app:results_breakdown} provides a detailed breakdown of the five error types for all interaction strategies.

Fig.~\ref{fig:mean_f1s} shows the evolution of mean F1 scores for visual grounding.
The three label-aware strategies obtain comparable grounding performance, whereas NoLabels+CaseMemory lags behind because novel part types are introduced without explicit labelling signals.
Crucially, the lower regret of Labels+TeacherRules is not explained by better visual grounding: its mean F1 is comparable to LabelsOnly and Labels+CaseMemory.
The gain therefore comes from how generic constraints are used by the symbolic planner to avoid semantically invalid assemblies, not merely from improved part classification.
Together, these results isolate the value of generic teacher statements as a direct channel for communicating reusable symbolic constraints during deployment.


\section{Related Work}

Our work relates to concept-based and explanatory interactive learning, where concept-level corrections, prototype interactions, and self-explanations revise neural or neurosymbolic models \citep{stammer2021right,stammer2022interactive,stammer2024learning}.
These approaches demonstrate the value of concept-level feedback, but typically focus on revising learned models with a largely fixed task vocabulary.
By contrast, our agent overcomes its unawareness of part predicates, subassembly concepts, topology, and semantic constraints during deployment, with the objective of immediate replanning rather than representation-level model debugging.

Recent VLM-based neurosymbolic systems map perception and language into symbolic structures such as predicates, visual programs, or planning constraints \citep{athalye2026pixels,wust2026synthesizing,kumar2026open}.
Human-guided systems such as LARA likewise integrate communication, learning, reasoning, and planning \citep{kokel2022lara}.
Our work studies a complementary regime: a human teacher communicates generic semantic constraints as verified symbolic knowledge during task execution, and an ASP planner immediately uses them for object-role assignment and join planning.
Closest to our own work are \citet{park2023interactive,park2025learning}, which use generic language for visually grounded concept and domain-theory learning; we extend this line to long-horizon assembly planning under quantified semantic constraints.


\section{Conclusion and Future Directions}

We presented a neurosymbolic agent that learns semantic assembly tasks through controlled natural language interaction and demonstrations.
The study focuses on deployment-time adaptation: the agent begins without the relevant part concepts, subassembly concepts, topology, or semantic constraints, and updates visual exemplars, symbolic memory, and (re)plans during task execution.
Our empirical findings show that teacher-provided generic constraints improve online assembly learning in this benchmark beyond part labels and non-generic correction memories alone, with gains arising from symbolic planning rather than visual grounding alone.
Together, these results illustrate a form of neurosymbolic adaptation in which explicit symbolic knowledge acquired during interaction immediately reshapes action selection over uncertain perceptual groundings.
The study remains limited to a controlled simulated domain and a modular, primarily perception-to-reasoning form of neurosymbolic coupling.
Future work should address more robust perception and language interfaces, cross-domain assembly benchmarks, abstraction from correction cases, and tighter symbolic-to-neural feedback where representation-level revision is needed.


\acks{
This work was supported by Informatics Global PhD Scholarships, funded by the School of Informatics at The University of Edinburgh.
Ramamoorthy is supported by a UKRI Turing AI World Leading Researcher Fellowship on AI for Person-Centred and Teachable Autonomy (grant EP/Z534833/1).
We thank the anonymous reviewers for their feedback on an earlier draft of this paper, and Rimvydas Rubavicius and Gautier Dagan for continued feedback over the course of this research.
Naver Labs Europe, the current employer of the attending author, sponsored registration and travel for in-person attendance to the conference venue.
}

\bibliography{nesy2026-sample}

\appendix

\section{ASP Encodings of Planning Subproblems}
\label{app:asp_encoding}

The ASP program given in Fig.~\ref{lst:asp_sp1_1}, \ref{lst:asp_sp1_2} and \ref{lst:asp_sp1_3} encodes the first planning subproblem described in the paper, i.e., goal selection.
Note that this ASP encoding assumes a solver that supports multi-shot solving feature like Clingo.

\begin{SaveVerbatim}{vrb:asp_sp1_1}
% Make choice for the top node (0), which represents the build target subassembly
1{ node_sa_template(0,S,I) : possible_root_template(S,I) }1.
% Determine possible template options for the root node, from the specified
% build target type and available supertype-subtype relations
possible_root_template(S,I) :- build_target(S), template_option(S,I).
possible_root_template(S2,I) :-
    build_target(S1), subtype_of(S1,S2), template_option(S2,I).

% Supertype-subtype relation is transitive
subtype_of(T1,T3) :- subtype_of(T1,T2), subtype_of(T2,T3).

% Annotating each child of a subassembly node N with its required atomic part
% concept type
node_atomic(n(N,NS),P) :- node_sa_template(N,S,I), req_atomic(S,I,NS,P).

% Meanwhile, a subassembly template may be directly specified for a node
node_sa_template(n(N,NS),S2,I2) :-
    node_sa_template(N,S1,I1), req_template(S1,I1,NS,S2,I2).

% Project to specify subassembly type of subassembly nodes
node_sa(N,S) :- node_sa_template(N,S,I).
node_sa(0,S) :- build_target(S).    % Also account for build target subtype
% Supertype-subtype info applied for substructures
node_sa(N,S2) :- node_sa(N,S1), subtype_of(S1,S2).

% Ancestor-descendant relations among nodes
component_node(n(N,NS),N) :- node_atomic(n(N,NS),_).
component_node(n(N,NS),N) :- node_sa(n(N,NS),_).
component_node(N1,N3) :- component_node(N1,N2), component_node(N2,N3).  % Transitive

% An object that is likely to be an instance of a type may fill an atomic node
% with matching type. Each object can fill up to one atomic node, and each
% atomic node can be filled by up to one object.
can_fill(O,N,P) :- node_atomic(N,P), type_likely(O,P,PR).
can_fill(O,N,P1) :- node_atomic(N,P2), subtype_of(P1,P2), type_likely(O,P1,PR).
{ fill_node(O,N) } :- can_fill(O,N,_), not must_unify(_,N).
:- type_likely(O,_,_), #count { N : fill_node(O,N) } > 1.
:- node_atomic(N,_), #count { O : fill_node(O,N) } > 1.

% Once determined an object to fill a specific atomic node, select exactly one
% part (sub)type to commit to among the visually licensed options 
1{ type_committed(O,P) : can_fill(O,N,P) }1 :- fill_node(O,N), can_fill(O,N,_).
% Supertype-subtype info percolates upwards
type_committed(O,P2) :- type_committed(O,P1), subtype_of(P1,P2).

% Always use objects labeled by user in response to agent's "Is there a ~?" queries
:- certified_label(O), not fill_node(O,_).
\end{SaveVerbatim}

\begin{SaveVerbatim}{vrb:asp_sp1_2}
% Identifying all objects selected to fill all descendant atomic nodes of a
% subassembly node
component_obj(O,N) :- node_sa(N,S), component_node(ND,N), fill_node(O,ND).

% For tracking which nodes will have to be connected at which contact points
to_connect(D1,D2,CP1,CP2) :-
    node_sa_template(N,S,I),
    fits_signature(D1,N,NS1,SG1), fits_signature(D2,N,NS2,SG2),
    connection_signature(S,I,NS1,NS2,SG1,SG2,CP1,CP2).
fits_signature(n(N,NS),N,NS,NS) :- node_atomic(n(N,NS),_).
fits_signature(D,N,NS,c(NS,SG)) :-
    fits_signature(D,n(N,NS),_,SG), node_sa_template(n(N,NS),_,_).

% Handling additional constraints due to already assembled parts. Atomic
% part type of existing object may or may not be specified, represented
% by arity of ext_obj predicate (ext_obj/2 vs. ext_obj/1). Similarly,
% contact sites between two joined existing parts may or may not be known
% (ext_conn/4 vs. ext_conn/2).
fresh_obj(O) :- component_obj(O,_), not ext_obj(O), not ext_obj(O,_).
    % 'Fresh' if not included in some subassembly

% If an existing object with known part type has exactly one atomic node with
% the matching type, unify
must_unify(O,N) :- ext_obj(O,P), 1{ node_atomic(_,P) }1, node_atomic(N,P).
% If a neighbor of a uniquely unified object connects with another by known
% contact site, unify (assumption here is that contact sites of each atomic part
% are all uniquely distinguishable)
must_unify(O1,N1) :- must_unify(O2,N2), ext_conn(O1,O2,CP1,CP2),
    to_connect(N1,N2,CP1,CP2).
% If a neighbor of a uniquely unified object has known type, and there exist
% exactly one node with matching type in the neighborhood of the uniquely unified
% node, unify
must_unify(O1,N1) :- must_unify(O2,N2), ext_obj(O1,P1), ext_conn(O1,O2),
    1{ node_atomic(N,P1) : to_connect(N,N2,_,_) }1,
    node_atomic(N1,P1), to_connect(N1,N2,_,_).
% If a neighbor of a uniquely unified object has known type, and there exist
% more than one nodes with matching type in the neighborhood of the uniquely
% unified node, *MAY* unify with one of them
may_unify(O1,N1) :- must_unify(O2,N2), ext_obj(O1,P1), ext_conn(O1,O2),
    2{ node_atomic(N,P1) : to_connect(N,N2,_,_) },
    node_atomic(N1,P1), to_connect(N1,N2,_,_).
% If one side of existing object connection uniquely unifies to an atomic node
% while the other's type is not specified, the latter *MAY* be unified with one
% of other atomic nodes connected to the unified node
may_unify(O1,N1) :- must_unify(O2,N2), ext_obj(O1), ext_conn(O1,O2),
    to_connect(N1,N2,_,_).
\end{SaveVerbatim}

\begin{SaveVerbatim}{vrb:asp_sp1_3}
% Observe must_unify relations
fill_node(O,N) :- must_unify(O,N).
% Different objects cannot be forced to unify with the same node
:- must_unify(O1,N1), must_unify(O2,N2), O1 != O2, N1 = N2.
% Observe ext_conn relations
:- ext_conn(O1,O2,_,_), fill_node(O1,N1), fill_node(O2,N2), not to_connect(N1,N2,_,_).
:- ext_conn(O1,O2), fill_node(O1,N1), fill_node(O2,N2), not to_connect(N1,N2,_,_).

% Supertype-subtype relations for ext_objs
ext_obj(O,P2) :- ext_obj(O,P1), subtype_of(P1,P2).

% Edge symmetricity
to_connect(N2,N1,CP2,CP1) :- to_connect(N1,N2,CP1,CP2).
ext_conn(O2,O1,CP2,CP1) :- ext_conn(O1,O2,CP1,CP2).
ext_conn(O2,O1) :- ext_conn(O1,O2).

% Compute average part compatibility score by taking average across all atomic
% nodes filled by recognized objects
node_score(N,PR) :-
    node_atomic(N,_), fill_node(O,N), type_committed(O,P), type_likely(O,P,PR).
avg_score(TS/NN) :- TS = #sum { NS,N : node_score(N,NS) },
    NN = #count { N : node_atomic(N,_) }, NN != 0.

% Prevent any violation of universally quantified constraints. Implemented as
% hard constraint, as violation can be evaded by deciding to not fill corresponding
% nodes with recognized objects.
:- forall_violation(_).

% Penalize any violation of existentially quantified constraints. Implemented
% as soft constraint, as plans violating these can still be admitted, where
% absence of required of parts will be handled by queries to user. In contrast,
% if implemented as hard constraints, many sane partial plans will be eliminated.
total_penalty(TP) :- TP = #sum { 30,EC : exists_violation(EC) }.

% Final goal configuration score
final_score(AS-TP) :- avg_score(AS), total_penalty(TP).

#program check(c).
#external query(c).
:- final_score(FS), query(c), FS <= c.
\end{SaveVerbatim}

\begin{figure}[p]
\floatconts
    {lst:asp_sp1_1}
    {\caption{ASP encoding of the goal selection subproblem: 1 of 3.}}
    {{\footnotesize \UseVerbatim{vrb:asp_sp1_1}}}
\end{figure}

\begin{figure}[p]
\floatconts
    {lst:asp_sp1_2}
    {\caption{ASP encoding of the goal selection subproblem: 2 of 3.}}
    {{\footnotesize \UseVerbatim{vrb:asp_sp1_2}}}
\end{figure}

\begin{figure}[p]
\floatconts
    {lst:asp_sp1_3}
    {\caption{ASP encoding of the goal selection subproblem: 3 of 3.}}
    {{\footnotesize \UseVerbatim{vrb:asp_sp1_3}}}
\end{figure}

The ASP program given in Fig.~\ref{lst:asp_sp2_1} and \ref{lst:asp_sp2_2} encodes the second planning subproblem, i.e., join sequence planning.
Note that this ASP encoding also assumes a solver that supports multi-shot solving feature, in addition to ASP-modulo-theory solving feature, again like Clingo.

\begin{SaveVerbatim}{vrb:asp_sp2_1}
%% Rules common to answer set planning
#program base.
holds(F,0) :- init(F).

#program step(t).
holds(F,t+1) :- holds(F,t), not -holds(F,t+1).
-holds(F,t+1) :- -holds(F,t), not holds(F,t+1).
1{ occ(A,t) : possible(A,t) }1.

#program check(t).
#external query(t).
:- query(t), not goal(t).

%% Rules common to (our way of encoding) assembly domains - Static laws
#program base.
% Part connection is a symmetric relation
to_connect(O2,O1) :- to_connect(O1,O2).

% Track every occasion where a subassembly formed in the previous timestep
% doesn't get used in the immediate next timestep (identifiable by index);
% total count of such occurrences will be minimized
#program step(t).
penalize(t) :-
    occ(join(S1,O1,S2,O2),t), holds(max_sa_index(S),t), S1 != S, S2 != S.

%% Rules common to assembly domains - Dynamic laws
#program step(t).
% join/4: Assembling two subassemblies at specified object; abstracts
% aligning & tightening.
% (Note: This implementation assumes we always assemble from right to left,
% without loss of generality)
possible(join(S1,O1,S2,O2),t) :-
    S1 != S2, O1 != O2,
    holds(part_of(O1,S1),t), holds(part_of(O2,S2),t),
    to_connect(O1,O2), not holds(connected(O1,O2),t).
holds(connected(O1,O2),t+1) :- occ(join(S1,O1,S2,O2),t).
holds(connected(O2,O1),t+1) :- occ(join(S1,O1,S2,O2),t).    % Symmetric
holds(max_sa_index(S+1),t+1) :- occ(join(S1,O1,S2,O2),t), holds(max_sa_index(S),t).
-holds(max_sa_index(S),t+1) :- occ(join(S1,O1,S2,O2),t), holds(max_sa_index(S),t).
-holds(part_of(O,S1),t+1) :- occ(join(S1,_,S2,_),t), holds(part_of(O,S1),t).
-holds(part_of(O,S2),t+1) :- occ(join(S1,_,S2,_),t), holds(part_of(O,S2),t).
holds(part_of(O,S+1),t+1) :-
    occ(join(S1,_,S2,_),t), holds(max_sa_index(S),t), holds(part_of(O,S1),t).
holds(part_of(O,S+1),t+1) :-
    occ(join(S1,_,S2,_),t), holds(max_sa_index(S),t), holds(part_of(O,S2),t).
\end{SaveVerbatim}

\begin{SaveVerbatim}{vrb:asp_sp2_2}
% Note: We don't consider 'disassemble' operation here; would need to treat
% subassembly predicate as fluent if we were to do this

% Minimize number of times occasions where agent does not immediately
% re-use the subassembly just assembled in the previous timestep for the
% next join action, thus increasing the number of primitive actions needed.
% It is important that this optimization statement is placed here, as part
% of the program fragment step(t).
#minimize { 1,T : penalize(T) }.

%% Goal condition; all object pairs to be connected have been 'connected'
% at time step t, either truly or speculatively
#program check(t).
goal(t) :- holds(connected(O1,O2),t) : to_connect(O1,O2).

%% Problem init; common to all problems
#program base.
% Initial configs
init(max_sa_index(-1)).
\end{SaveVerbatim}

\begin{figure}[p]
\floatconts
    {lst:asp_sp2_1}
    {\caption{ASP encoding of the join sequence planning subproblem: 1 of 2.}}
    {{\footnotesize \UseVerbatim{vrb:asp_sp2_1}}}
\end{figure}

\begin{figure}[t]
\floatconts
    {lst:asp_sp2_2}
    {\caption{ASP encoding of the join sequence planning subproblem: 2 of 2.}}
    {{\footnotesize \UseVerbatim{vrb:asp_sp2_2}}}
\end{figure}

\section{Detailed Account of Agent-Teacher Interaction Flow}
\label{app:interaction_flow}

This section expands the compact description of the controlled interaction protocol in \S\ref{sec:semantic_assembly:user_interactions}.
Fig.~\ref{fig:interaction_flow} depicts the full set of teacher-learner interaction branches used in the experiments.

\begin{figure}[t]
\floatconts
    {fig:interaction_flow}
    {\caption{Possible teacher-learner interactions. $\square$ marks termination of an episode.}}
    {\includegraphics[width=0.825\textwidth]{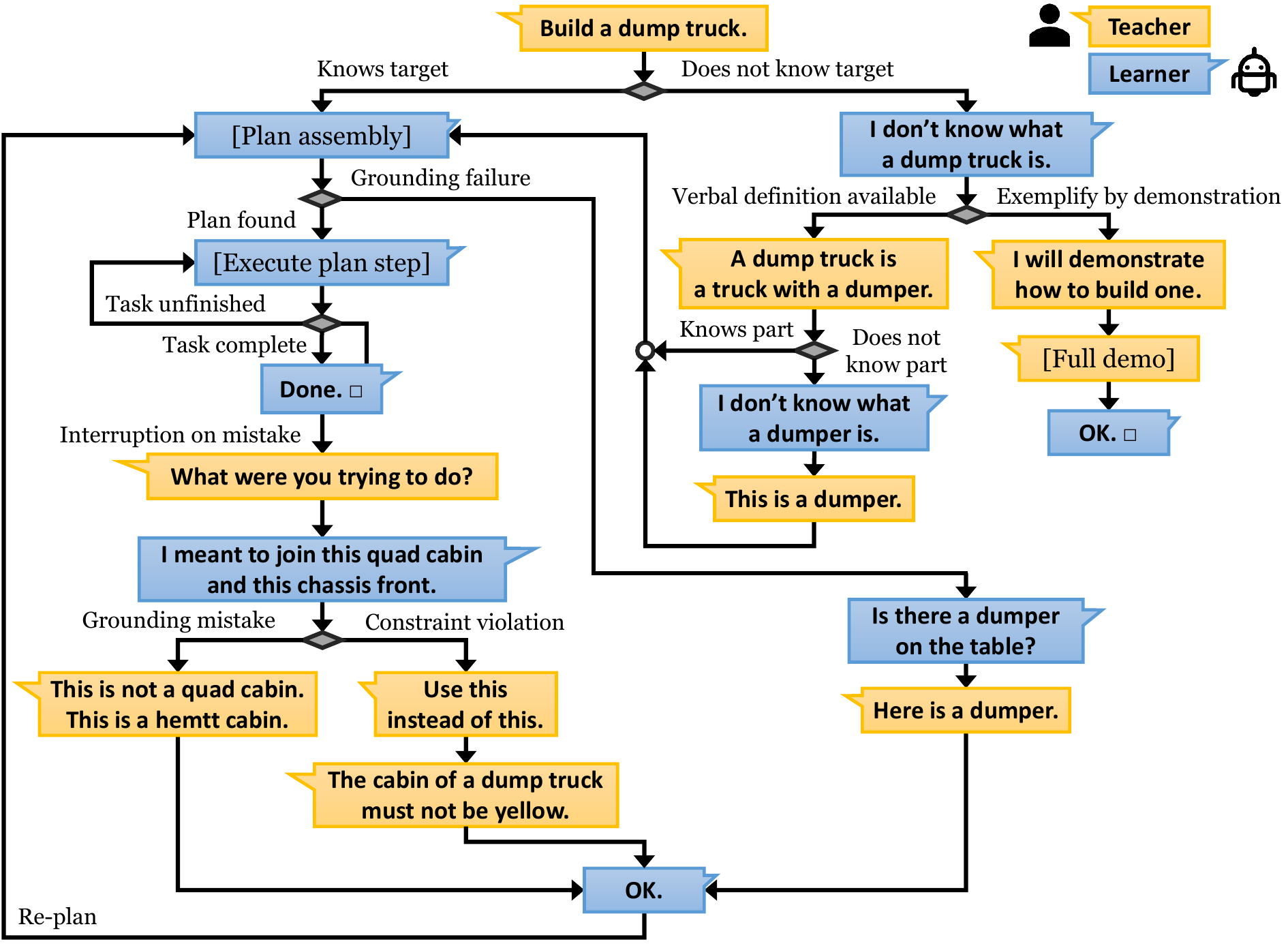}}
\end{figure}

Each task begins with the teacher specifying the assembly goal.
If the agent is unaware of the assembly target, it reports its ignorance and the teacher responds either through a full demonstration or through a verbal definition, depending on context.
Demonstrations (accompanied with NL narrations) are preferred when new geometric relations or join poses must be learned.
Verbal definitions suffice if the agent already possesses the relevant intermediary concepts (e.g., ``A dump truck is a truck with a dumper").

If the agent believes it has sufficient knowledge, it generates a plan and begins execution.
During planning, the agent may fail to visually recognise a required atomic part (which is guaranteed to exist in $C$).
In such cases, the agent requests clarification and the teacher responds by pointing to the correct object.

If the agent makes an error during execution, the teacher interrupts and asks the agent to explain its intended action.
The agent's response reveals its current beliefs about the involved objects, and exposes two possible knowledge gaps.
First, the agent may have made a grounding error.
The teacher corrects this by negating the incorrect grounding and asserting the correct type.
Second, the grounding may be correct but the attempted action may violate a semantic constraint.
The teacher then points to a valid object and expresses the violated constraint as a generic NL statement.
After corrective feedback, the agent updates its knowledge and replans.

\section{Vision Processing Module: Implementation details}
\label{app:vision_module}

The vision processing module parses the agent's visual inputs from its RGB sensor into structured representations that other modules can further process.
The module deals with three visual subtasks: object detection (including predictions of its types); point cloud extraction; and pose estimation.

The goal of object detection is to localise and to classify scene objects.
We use binary segmentation masks to represent localised object instances instead of bounding boxes since they allow more accurate visual feature extraction for classification.
Our primary focus is on learning to recognise classes of objects rather than their locations, since this directly interfaces with the goal of identifying novel parts.
While high-quality, off-the-shelf segmentation models such as SAM \citep{kirillov2023segment} could be used for object localization, it would introduce uncontrolled confounders for testing our main hypotheses.
As stated in \S\ref{sec:semantic_assembly:assumptions}, experiments use ground-truth masks obtained from the simulated environment.
 
We implement few-shot open-set classification by keeping a collection of lightweight binary classifiers for each atomic part concept that the learner is currently aware of.
(This subset of `aware' part concepts $\Pi$ keeps expanding as interaction with the teacher proceeds.)
The binary classifiers are induced from positive and negative exemplars stored in the agent's long-term memory and gained from teacher interactions (see \S\ref{sec:semantic_assembly:user_interactions}).
Our implementation uses the DINOv2-base model \citep{oquab2023dinov2} as a feature extractor backbone, which takes a scene image and object masks as input, and outputs are fed into RBF-kernel SVMs to yield per-concept predictions.

Atomic part concepts need to have proper representations of their 3D structures for accurate manipulation during assembly.
Under the type-level geometry assumption stated in \S\ref{sec:semantic_assembly:assumptions}, each atomic part concept is represented by a point cloud.
When the learner encounters an atomic part type for the first time, and notices that it does not correspond to any previously known types, the learner registers a new atomic part concept in $\Pi$ and associates it with a point cloud representation extracted from the instance.
Given that visual inputs are single-view RGB images, the learner picks up the novel part concept instance and collects a set of images of the object observed from different viewpoints with known poses.
The feature extractor processes each image to obtain per-pixel embeddings at a lower resolution, which can be used as point descriptors to be fed into a photogrammetric point cloud extraction algorithm.
Our implementation employs COLMAP \citep{schonberger2016structure} for point cloud reconstruction.

Estimation of 6D poses of objects is necessary to achieve accurate joining of parts at specified relative poses.
Few-shot or zero-shot object pose estimation with RGB images has witnessed significant research progress with the advent of powerful vision foundation models \citep{liu2022gen6d, ornek2024foundpose}.
As stated in \S\ref{sec:semantic_assembly:assumptions}, experiments use ground-truth object poses from the simulated environment after each manipulation action.

\section{Per-strategy Details on Learning from Visual Data}
\label{app:learning_vis}

\subsection{Learning Visual Grounding of Parts}

The agent's visual grounding performance is determined by the binary classifiers induced from the positive and negative exemplar sets $\chi^{+/-}_\pi$ stored in the visual XB.
The progress of learning to visually recognise atomic parts $\Pi$ largely depends on acquiring labelled concept exemplars through interactions with the teacher.
Consequently, the visual grounding learning process differs between NoLabels+CaseMemory and the label-aware strategies, since only the latter have access to NL labels for concepts in $\Pi$.

Label-aware agents directly obtain exemplar labelling through concept labelling statements like ``This is a \textsf{dumper}", ``This is not a \textsf{quad\_cabin}", ``Here is a \textsf{large\_wheel}".
As illustrated in Fig.~\ref{fig:interaction_flow}, these labelling statements are given after the agent's utterance that reveals how it has mistaken a part type for another (``I meant to join this X and this Y.") or failed to ground a needed instance (``Is there a X on the table?").
Concept labels are also provided when the teacher provides full demonstrations upon the agent's ignorance of goal subassembly concepts, in which each action step is narrated in natural language.
These interactions are possible only when the agent shares a NL vocabulary with the teacher.

In contrast, NoLabels+CaseMemory agents are neither able to understand the teacher's concept labelling statements nor to generate utterances that expose its imperfect grounding capability due to lack of a shared vocabulary.
In case of grounding failures, the agent cannot ask ``Is there a X on the table?" and has to report ``I cannot find a part I need on the table" instead, after which the teacher will simply demonstrate a valid part join.
After agent mistakes, the teacher cannot ask the probing question ``What were you trying to join?" because agent cannot answer with its current part recognitions, so the agent is simply interrupted; afterwards, the agent will immediately request a demonstration of a valid part join.

Accordingly, the NoLabels+CaseMemory baseline in our experiments has to rely on indirect learning signals to update their exemplar sets: post-hoc analysis of each completed episode, and the teacher's interruptions upon grounding mistakes.
Post-hoc analysis of an episode updates positive exemplar sets $\chi_\pi^+$ by matching the topology of the completed structure against $G_\gamma$ via graph matching.
Negative exemplars can be inferred from teacher interruptions, which can serve as `pairwise negative labels' of the involved object pairs.
Specifically, if the join of objects $o_1$ and $o_2$ are interrupted as invalid by the teacher due to incorrect grounding, and the agent had recognised $o_1,o_2$ as $\pi_1,\pi_2\in\Pi$ respectively, then it is not the case that $o_1$ is a $\pi_1$ and $o_2$ is a $\pi_2$ at the same time; otherwise, the join would not have been interrupted.
From the interruption, if the agent is certain that $o_1$ is indeed an instance of $\pi_1$, it can logically entail $o_2$ is not an instance of $\pi_2$, thereby updating $\chi_{\pi_2}^-$.
Note that both label acquisition processes are susceptible to noise.

\subsection{Learning 3D Geometries of Novel Parts and Joins}

Across the four interaction strategies, the process itself remains the same for learning point cloud representations of atomic parts and joining poses.
The practical difference between NoLabels+CaseMemory vs. label-aware agents in this regard has to do with \textit{when} the point cloud extraction procedure is triggered during interactions.
As explained in \S\ref{sec:architecture:vision_processing}, the point cloud for an atomic part type is extracted when the agent encounters its instance for the first time in its operation.
In other words, the agent must first become aware that it has encountered a novel part type that is not in $\Pi$ prior to extracting its 3D geometry.
This is straightforward for label-aware agents, who can immediately recognise when the teacher introduces a novel part type by use of neologisms.

Conversely, NoLabels+CaseMemory agents have to rely on their own judgements to determine whether an object is an instance of a novel atomic part type.
In our implementation, NoLabels+CaseMemory agents register a novel part type when an object used in the teacher's full demonstration obtains low likelihood scores from few-shot visual recognition across all currently known types in $\Pi$.
Join poses involving the novel type are then obtained from the teacher's demonstration.
Note that NoLabels+CaseMemory agents may mistakenly identify a familiar part type as novel, or conversely, fail to recognise a novel part as unfamiliar, which will likely result in downstream errors during task execution.

\section{Full Description of Toy Truck Assembly Domain}
\label{app:domain_full}

Our simulated toy truck domain comprises 22 atomic part types in $\Pi$.
See Tab.~\ref{tab:parts} for the full list of the part types, whether they can have colour attributes, and example visuals.

\begin{table}[p]
\floatconts
    {tab:parts}
    {\caption{Atomic part types in our simulated toy truck domain. Coloured parts may have one of five colours: red, green, blue, yellow, white.}}
    {
        \begin{adjustbox}{width=0.45\textwidth}
        \begin{tabular}{p{1.4in}|p{0.4in}|M{0.7in}}
        \toprule
        \multicolumn{1}{c|}{Name} & \multicolumn{1}{c|}{Coloured} & \multicolumn{1}{c}{Sample image} \\
        \midrule
        \textsf{quad\_cabin} & True & \includegraphics[height=0.6in]{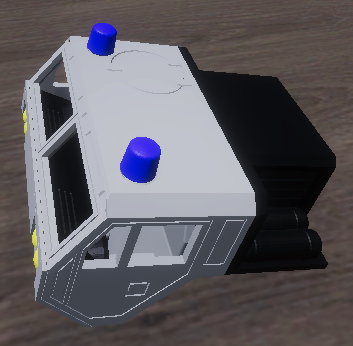} \\ 
        \textsf{hemtt\_cabin} & True & \includegraphics[height=0.6in]{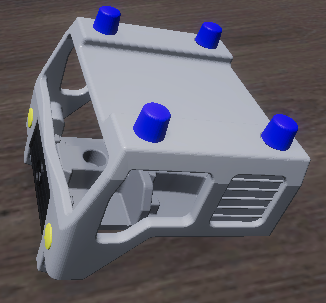} \\ 
        \textsf{chassis\_front} & False & \includegraphics[height=0.6in]{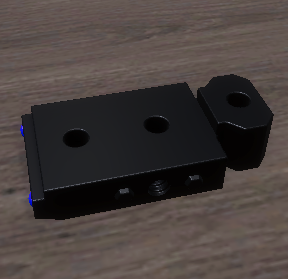} \\ 
        \textsf{chassis\_back} & False & \includegraphics[height=0.6in]{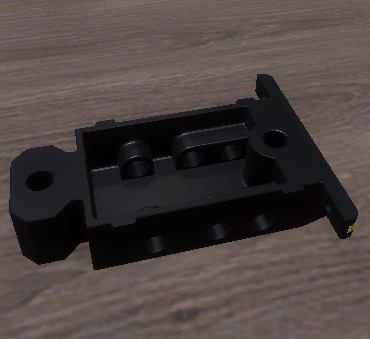} \\
        \textsf{flat\_chassis\_center} & True & \includegraphics[height=0.6in]{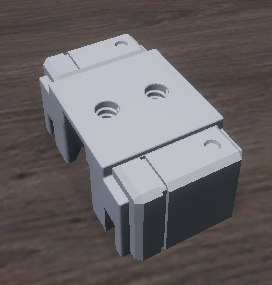} \\ 
        \textsf{spares\_chassis\_center} & True & \includegraphics[height=0.6in]{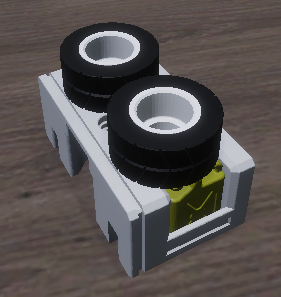} \\ 
        \textsf{staircase\_chassis\_center} & True & \includegraphics[height=0.6in]{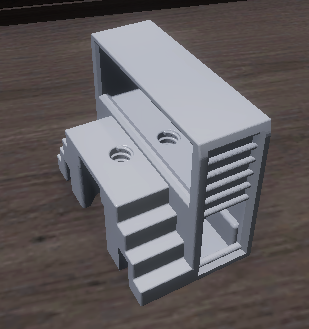} \\ 
        \textsf{platform} & False & \includegraphics[height=0.6in]{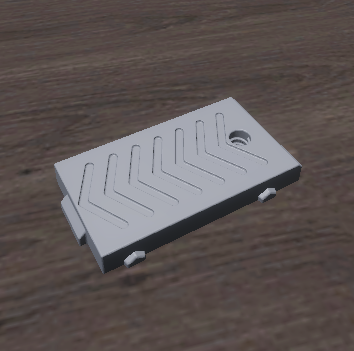} \\
        \textsf{dumper} & False & \includegraphics[height=0.6in]{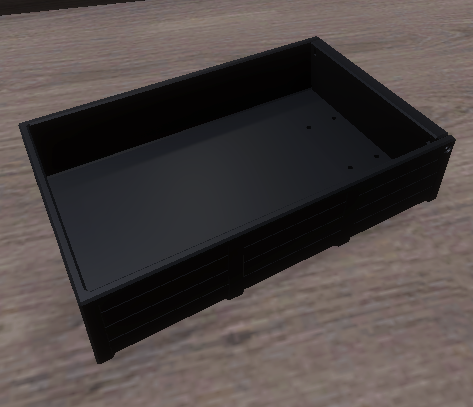} \\
        \textsf{ladder} & False & \includegraphics[height=0.6in]{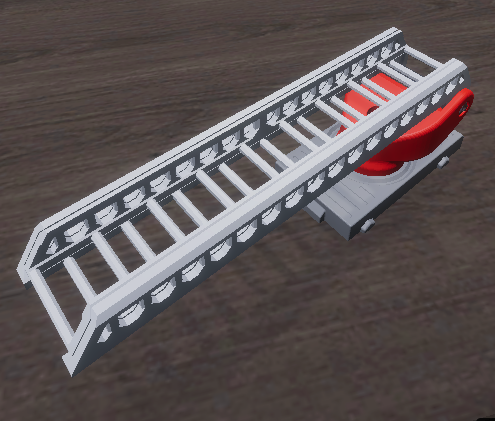} \\
        \textsf{rocket\_launcher} & False & \includegraphics[height=0.6in]{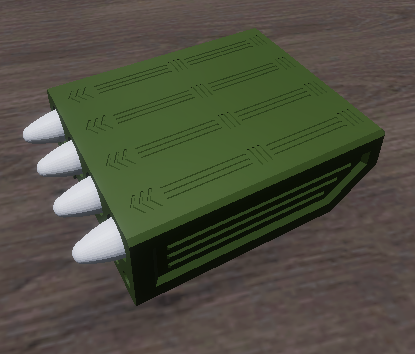} \\
        \bottomrule
        \end{tabular}
        \end{adjustbox}
        \quad
        \begin{adjustbox}{width=0.45\textwidth}
        \begin{tabular}{p{1.4in}|p{0.4in}|M{0.7in}}
        \toprule
        \multicolumn{1}{c|}{Name} & \multicolumn{1}{c|}{Coloured} & \multicolumn{1}{c}{Sample image} \\
        \midrule
        \textsf{normal\_fl\_fender} & True & \includegraphics[height=0.6in]{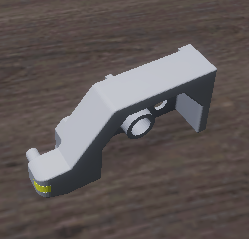} \\ 
        \textsf{normal\_fr\_fender} & True & \includegraphics[height=0.6in]{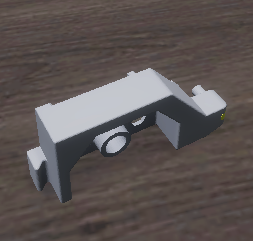} \\ 
        \textsf{normal\_bl\_fender} & True & \includegraphics[height=0.6in]{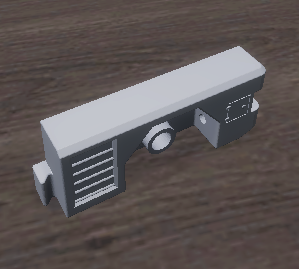} \\ 
        \textsf{normal\_br\_fender} & True & \includegraphics[height=0.6in]{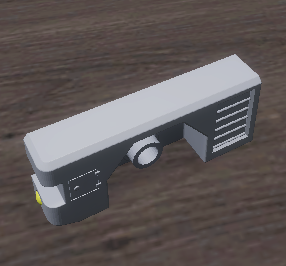} \\
        \textsf{large\_fl\_fender} & True & \includegraphics[height=0.6in]{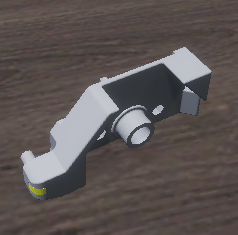} \\ 
        \textsf{large\_fr\_fender} & True & \includegraphics[height=0.6in]{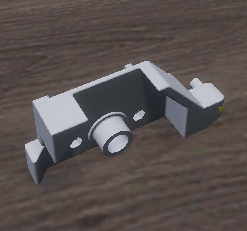} \\ 
        \textsf{large\_bl\_fender} & True & \includegraphics[height=0.6in]{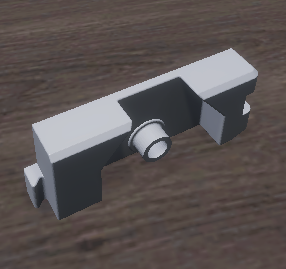} \\ 
        \textsf{large\_br\_fender} & True & \includegraphics[height=0.6in]{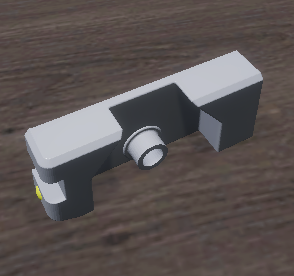} \\
        \textsf{normal\_wheel} & False & \includegraphics[height=0.6in]{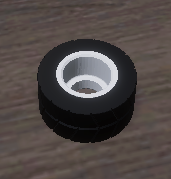} \\
        \textsf{large\_wheel} & False & \includegraphics[height=0.6in]{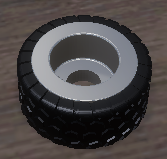} \\
        \textsf{bolt} & False & \includegraphics[height=0.6in]{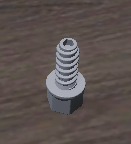} \\
        \bottomrule
        \end{tabular}
        \end{adjustbox}
    }
\end{table}

\noindent
$\Pi$ also includes their supertypes:
\begin{itemize}
    \item \{\textsf{cabin}, \textsf{load} \textsf{chassis\_center}, \textsf{fender}, \textsf{wheel}, \textsf{fl\_fender}, \textsf{fr\_fender}, \textsf{bl\_fender}, \textsf{br\_fender}, \textsf{normal\_fender}, \textsf{large\_fender}\}.
\end{itemize}
See the domain theory $\Omega$ for hyper/hyponymy relations that hold among them.

\noindent
The set of subassembly concepts $\Sigma$ include goal concepts, which are specified as task goals, and intermediary concepts, which are semantically meaningful substructures of goal concepts.
\begin{itemize}
    \item Goal subassemblies: \{\textsf{truck}, \textsf{fire\_truck}, \textsf{dump\_truck}, \textsf{container\_truck}, \textsf{missile\_truck}\}.
    \item Non-goal subeassemblies: \{\textsf{truck\_front}, \textsf{truck\_back}, \textsf{fl\_fw\_unit}, \textsf{fr\_fw\_unit}, \textsf{bl\_fw\_unit}, \textsf{br\_fw\_unit}, \textsf{normal\_fl\_fw\_unit}, \textsf{normal\_fr\_fw\_unit}, \textsf{normal\_bl\_fw\_unit}, \textsf{normal\_br\_fw\_unit},\\ \textsf{large\_fl\_fw\_unit}, \textsf{large\_fr\_fw\_unit}, \textsf{large\_bl\_fw\_unit}, \textsf{large\_br\_fw\_unit}\}.
\end{itemize}
Likewise, see the domain theory $\Omega$ for hyper/hyponymy relations that hold among them.
See the assembly topology graphs $G_\gamma$ in Fig.~\ref{fig:topology} for the makeups of the subassembly concepts.

The domain theory $\Omega$ features hyper/hyponymy relations and holo/meronymy relations.
Fig.~\ref{fig:taxonomy} depicts all hyper/hyponymy relations among concepts in $\Pi$ and $\Sigma$.
Holo/meronymy relations can be induced from the assembly topology graphs $G_\gamma$ in Fig.~\ref{fig:topology}.

\begin{figure}[h]
\floatconts
    {fig:taxonomy}
    {\caption{Supertype-subtype relations in our simulated truck domain.}}
    {\includegraphics[width=\textwidth]{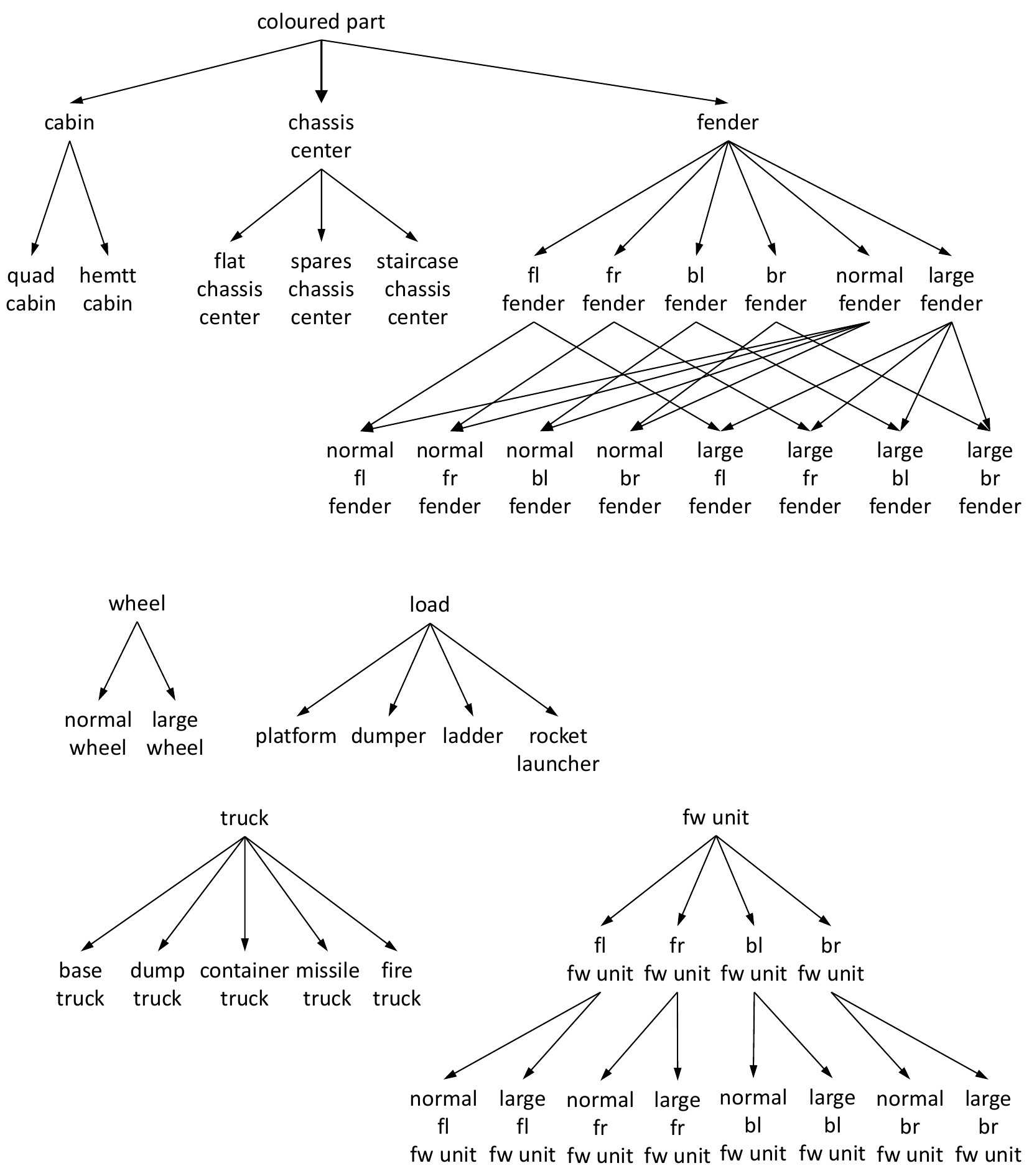}}
\end{figure}

The assembly topology graphs $G_\gamma$ describe how each subassembly in $\Sigma$ is composed from atomic parts in $\Pi$ or other subassemblies in $\Sigma$.
$G_\gamma$ for the truck subtypes $\gamma\in$\{\textsf{fire\_truck}, \textsf{dump\_truck}, \textsf{container\_truck}, \textsf{missile\_truck}\} inherit from $G_\text{\textsf{truck}}$; they are further defined with their definitional semantic constraints in $S_\gamma$.
Fig.~\ref{fig:topology_flat} illustrates how the `fully flattened' topology of a \textsf{truck} looks like.

\begin{figure}[htbp]
\floatconts
    {fig:topology}
    {\caption{Assembly topology graphs $G_\gamma$ for each $\gamma\in\Sigma$. Blue circle nodes denote atomic parts, whereas grey square nodes denote subassembly parts. $G_\gamma$ for $\gamma\in$\{\textsf{normal\_fr\_fw\_unit}, \textsf{normal\_bl\_fw\_unit}, \textsf{normal\_br\_fw\_unit}, \textsf{large\_fl\_fw\_unit}, \textsf{large\_fr\_fw\_unit}, \textsf{large\_bl\_fw\_unit}, \textsf{large\_br\_fw\_unit}\} are parallel to the graph for \textsf{normal\_fl\_fw\_unit} except the fender piece, so they are omitted as redundant.}}
    {\includegraphics[width=0.8\textwidth]{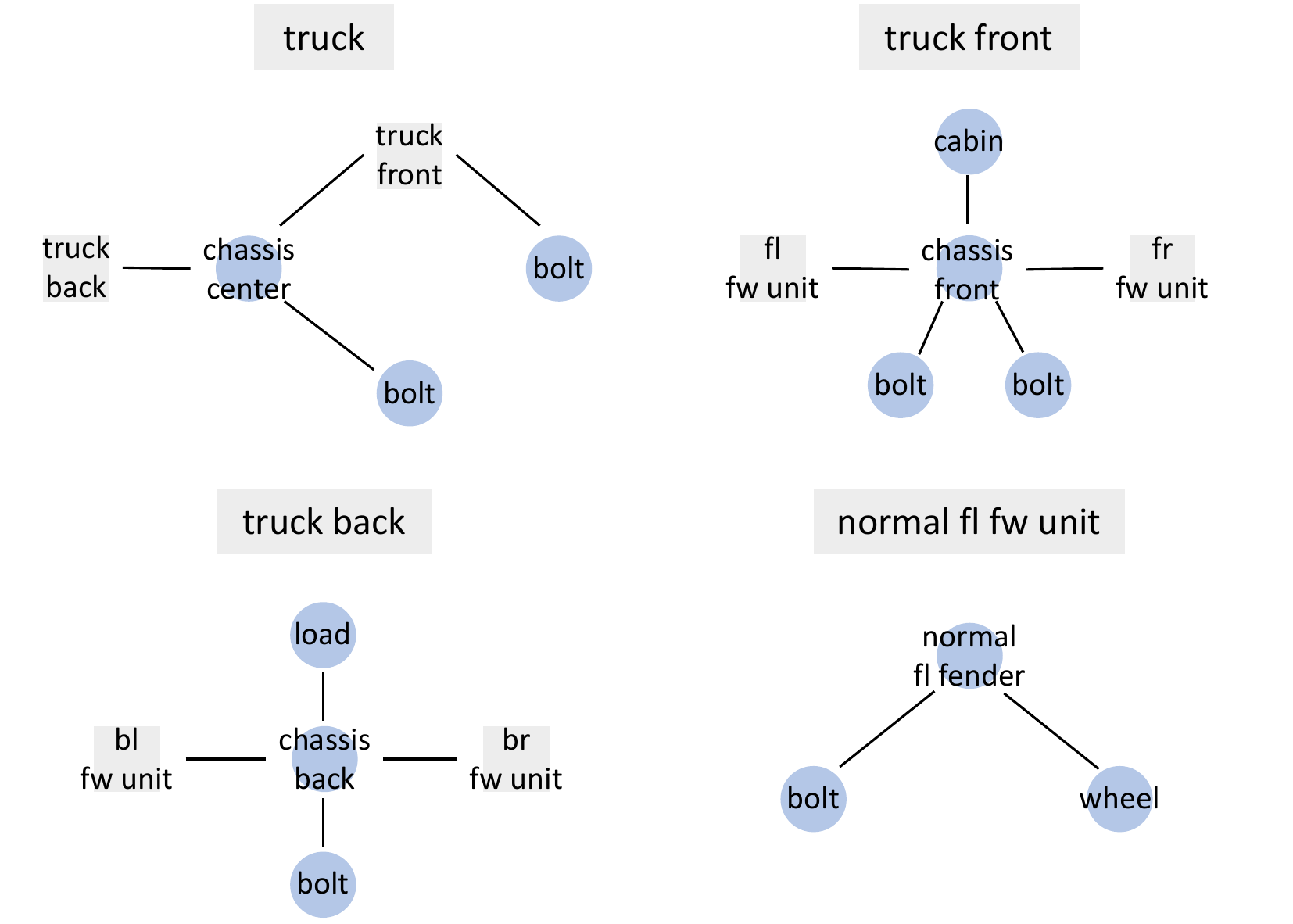}}
\end{figure}

\begin{figure}[htbp]
\floatconts
    {fig:topology_flat}
    {\caption{$G_\text{\textsf{truck}}$, the assembly topology of \textsf{truck}, `flattened out' by recursively replacing subassembly nodes with their respective $G_\gamma$.}}
    {\includegraphics[width=0.55\textwidth]{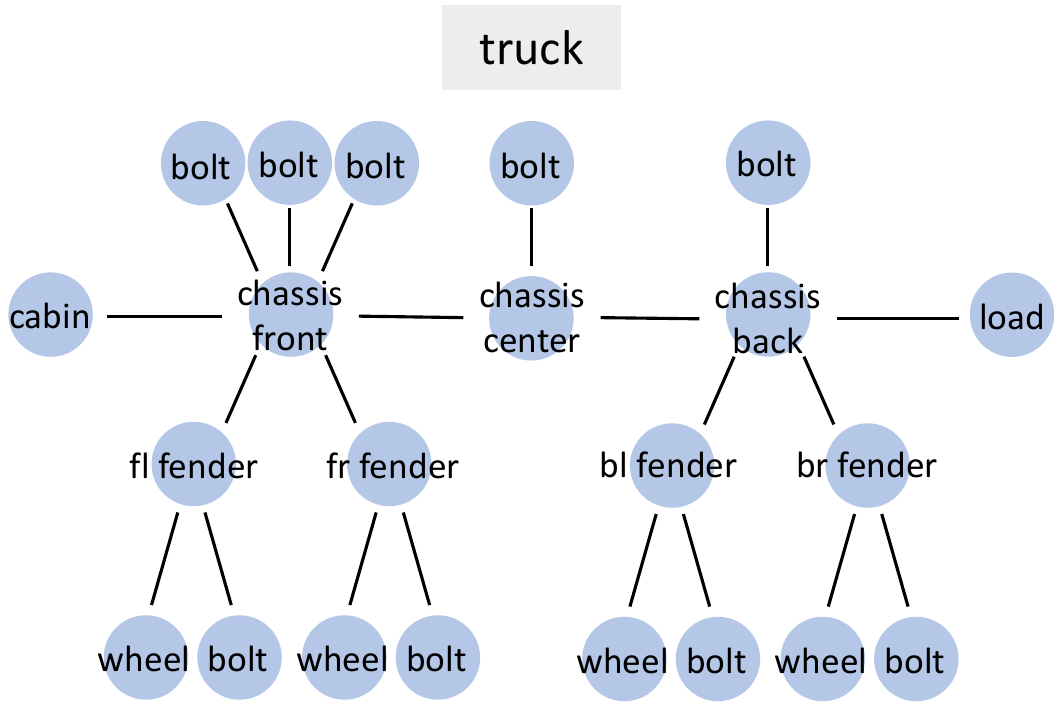}}
\end{figure}

Finally, Tab.~\ref{tab:constraints} provides the full list of the semantic constraints $S_\gamma$ for $\gamma\in\Sigma$, merged into a single table.
Note how many of the constraints, or their more specific variants, cannot be obtained by the most-specific correction memories used by the +CaseMemory strategies.

\begin{table}[p]
\floatconts
    {tab:constraints}
    {\caption{The set of all semantic constraints in our toy truck assembly domain. $\bot$ denotes falsity, which must not be entailed.}}
    {
        \begin{adjustbox}{width=\textwidth}
        \begin{tabular}{l|l}\toprule
            \multicolumn{1}{c|}{NL generic statement} & \multicolumn{1}{c}{FOL encoding} \\ \midrule
            ``A base truck has a platform." & $\forall x\exists y.baseTruck(x)\rightarrow platform(y)\land have(x,y)$ \\ \midrule
            \multirow{2}{2.8in}{``A dump truck has a dumper and a quad cabin."} & $\forall x\exists y\exists z.dumpTruck(x)\rightarrow$ \\
            & \qquad$dumper(y)\land have(x,y)\land quadCabin(z)\land have(x,z)$ \\ \midrule
            \multirow{2}{2.8in}{``A container truck has a dumper and a hemtt cabin."} & $\forall x\exists y\exists z.containerTruck(x)\rightarrow$ \\
            & \qquad$dumper(y)\land have(x,y)\land hemttCabin(z)\land have(x,z)$ \\ \midrule
            ``A missile truck has a rocket launcher." & $\forall x\exists y.missileTruck(x)\rightarrow rocketLauncher(y)\land have(x,y)$ \\ \midrule
            \multirow{2}{2.8in}{``A fire truck has a ladder and a staircase center."} & $\forall x\exists y\exists z.fireTruck(x)\rightarrow$ \\
            & \qquad$ladder(y)\land have(x,y)\land staircaseCenter(z)\land have(x,z)$ \\ \midrule
            \multirow{2}{2.8in}{``All fender pairs of a truck front must have same color."} & $\forall x\forall y\forall z.truckFront(x)\land fender(y)\land fender(z)\land$ \\
            & \qquad $have(x,y),have(x,z)\rightarrow sameColor(y,z)$ \\ \midrule
            \multirow{2}{2.8in}{``All fender pairs of a truck back must have same color."} & $\forall x\forall y\forall z.truckBack(x)\land fender(y)\land fender(z)\land$ \\
            & \qquad $have(x,y),have(x,z)\rightarrow sameColor(y,z)$ \\ \midrule
            \multirow{2}{2.8in}{``A fw-unit must not have a large fender and a normal wheel."} & $\forall x\forall y\forall z.fwUnit(x)\land largeFender(y)\land normalWheel(z)\land$ \\
            & \qquad $have(x,y),have(x,z)\rightarrow\bot$ \\ \midrule
            \multirow{2}{2.8in}{``A fw-unit must not have a normal fender and a large wheel."} & $\forall x\forall y\forall z.fwUnit(x)\land normalFender(y)\land largeWheel(z)\land$ \\
            & \qquad $have(x,y),have(x,z)\rightarrow\bot$ \\ \midrule
            \multirow{2}{2.8in}{``All fender-center pairs of a truck must not have same color."} & $\forall x\forall y\forall z.truck(x)\land fender(y)\land chassisCenter(z)\land$ \\
            & \qquad $have(x,y),have(x,z)\rightarrow\neg sameColor(y,z)$ \\ \midrule
            \multirow{2}{2.8in}{``A truck must not have a normal wheel and a large wheel."} & $\forall x \forall y \forall z.truck(x)\land normalWheel(y)\land largeWheel(z)\land$ \\ 
            & \qquad $have(x,y),have(x,z)\rightarrow\bot$ \\ \midrule
            \multirow{2}{2.8in}{``All cabin-center pairs of a base truck must not have same color."} & $\forall x\forall y\forall z.baseTruck(x)\land cabin(y)\land chassisCenter(z)\land$ \\
            & \qquad $have(x,y),have(x,z)\rightarrow\neg sameColor(y,z)$ \\ \midrule
            ``A base truck must not have a staircase center." & $\forall x \forall y.baseTruck(x)\land staircaseCenter(y)\land have(x,y)\rightarrow\bot$ \\ \midrule
            ``A missile truck has a green fender." & $\forall x\exists y.missileTruck(x) \rightarrow fender(y)\land green(y)\land have(x,y)$ \\ \midrule
            \multirow{2}{2.8in}{``All cabin-center pairs of a missile truck must have same color."} & $\forall x\forall y\forall z.missileTruck(x)\land cabin(y)\land chassisCenter(z)\land$ \\
            & \qquad $have(x,y),have(x,z)\rightarrow sameColor(y,z)$ \\ \midrule
            \multirow{2}{2.8in}{``All coloured parts of a dump truck must not be yellow."} & $\forall x\forall y.dumpTruck(x)\land coloredPart(y)\land have(x,y)\rightarrow$ \\
            & \qquad$\neg yellow(y)$ \\ \midrule
            \multirow{2}{2.8in}{``All wheels of a dump truck must be normal wheels."} & $\forall x\forall y.dumpTruck(x)\land wheel(y)\land have(x,y)\rightarrow$ \\
            & \qquad$normalWheel(y)$ \\ \midrule
            \multirow{2}{2.8in}{``All coloured parts of a container truck must not be blue."} & $\forall x\forall y.containerTruck(x)\land coloredPart(y)\land have(x,y)\rightarrow$ \\
            & \qquad$\neg blue(y)$ \\ \midrule
            \multirow{2}{2.8in}{``All wheels of a container truck must be large wheels."} & $\forall x\forall y.containerTruck(x)\land wheel(y)\land have(x,y)\rightarrow$ \\
            & \qquad$largeWheel(y)$ \\ \midrule
            ``A fire truck has a white coloured part." & $\forall x\exists y.fireTruck(x) \rightarrow coloredPart(y)\land white(y)\land have(x,y)$ \\ \midrule
            ``All fender of a fire truck must be red." & $\forall x\forall y.fireTruck(x)\land fender(y)\land have(x,y)\rightarrow red(y)$\\ \bottomrule
        \end{tabular}
        \end{adjustbox}
    }
\end{table}

\section{Breakdown of Cumulative Regrets}
\label{app:results_breakdown}

Fig.~\ref{fig:error_types} provides a detailed breakdown of the averaged cumulative regret curves reported in the main paper.

\begin{figure}[h]
\floatconts
    {fig:error_types}
    {\caption{Breakdown of the cumulative regrets into the five agent error types.}}
    {
        \subfigure[Structurally incompatible join attempt]{
            \includegraphics[width=0.475\textwidth]{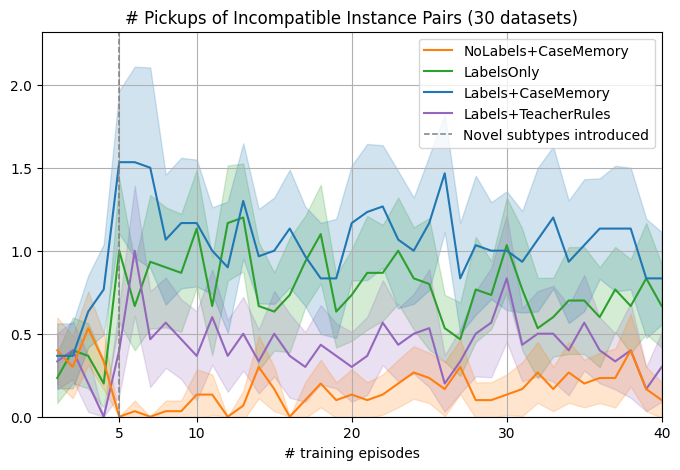}
        }
        \subfigure[Join at incorrect pose]{
            \includegraphics[width=0.475\textwidth]{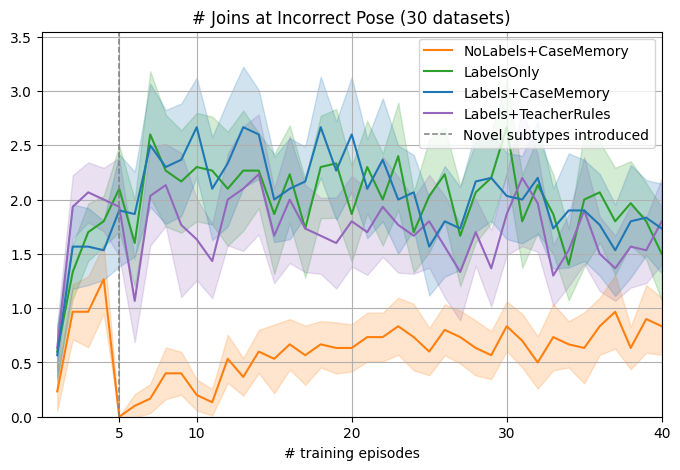}
        }
        \subfigure[Distractor usage]{
            \includegraphics[width=0.475\textwidth]{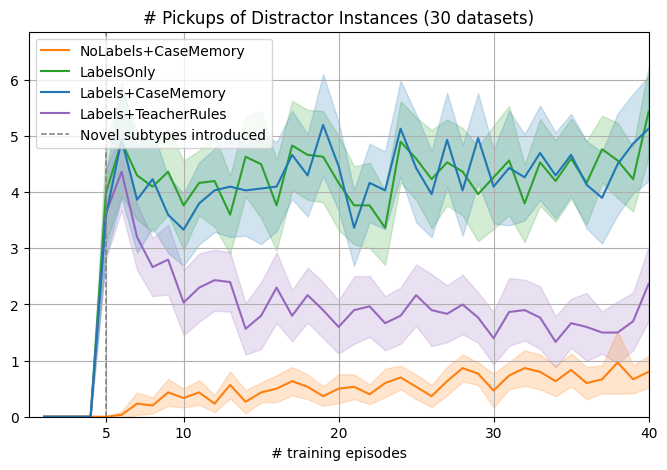}
        }
        \subfigure[Grounding failure]{
        \label{fig:topology:fw_unit}
            \includegraphics[width=0.475\textwidth]{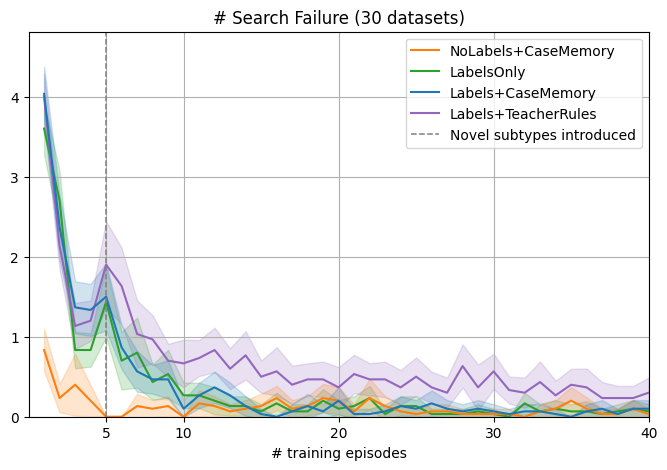}
        }
        \subfigure[Planner timeout]{
            \includegraphics[width=0.475\textwidth]{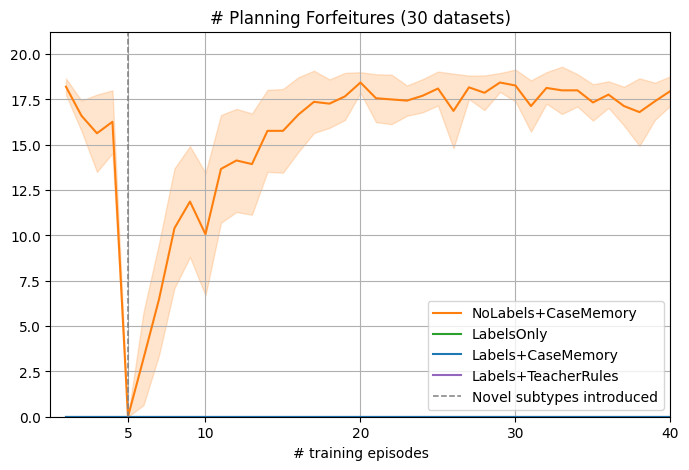}
        }
    }
\end{figure}

Note in particular how Labels+TeacherRules agents make fewer distractor-usage mistakes as learning proceeds.
The idiosyncratic behaviours of NoLabels+CaseMemory curves, namely the seemingly increasing error counts for most types after the warm-up problems, are due to the fact that they need full demonstrations for novel truck types, unlike label-aware agents for which verbal definitions suffice; such demonstrations do not count towards any error type.
It is randomly determined at which point novel truck types are introduced in each dataset, so the numbers are averaged out across the datasets, giving the false sense of `increasing errors' as the learning progresses.
Indeed, after sufficient number of episodes, NoLabels+CaseMemory agents start to show learning progress again for most error types.

\section{Auxiliary Metric: Unique Motion Planner Calls}
\label{app:results_aux}

Fig.~\ref{fig:collision_checks} provides the accumulated unique calls to the motion planner integrated into ASP planning procedure, invoked for collision-free joining path checks.

\begin{figure}[t]
\floatconts
    {fig:collision_checks}
    {\caption{Accumulated counts of unique calls to the integrated motion planner.}}
    {\includegraphics[width=0.75\textwidth]{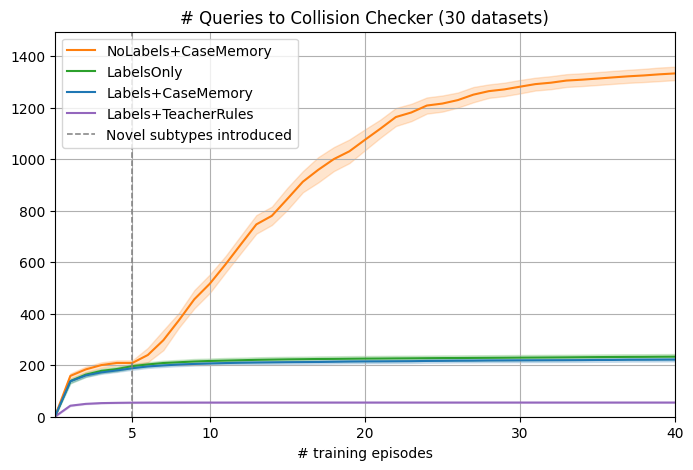}}
\end{figure}

NoLabels+CaseMemory makes substantially more motion-planner calls than the label-aware strategies, largely because poorer grounding produces more invalid object choices and hence more replanning.
Among the label-aware strategies, Labels+TeacherRules makes fewer calls because semantic constraints are incorporated already in the goal-selection ASP subproblem: object-role assignments that would violate known constraints are rejected or penalised before the join-sequence planner invokes collision checks.
This can reduce calls to the motion planner when semantically invalid distractor choices would otherwise remain geometrically feasible.
Since this metric is auxiliary and sensitive to implementation details of the planner and collision checker, we do not use it as primary evidence for the semantic-learning claim.





\end{document}